%% file: main.tex
\documentclass[conference,a4paper]{IEEEtran}
\usepackage[left=1.62cm,right=1.62cm,top=1.9cm]{geometry}

\IEEEoverridecommandlockouts

\usepackage{graphicx}
\usepackage{subfigure}
\usepackage{tabularx}
\usepackage{booktabs}
\usepackage{multirow}
\usepackage{siunitx}
\usepackage{xcolor}
\usepackage{algorithm}
\usepackage{comment}
\usepackage{todonotes}
\usepackage{varwidth}
\usepackage{subfig} 
\usepackage{algpseudocode}
\usepackage{enumitem}
\usepackage{caption}
\usepackage{balance}
\usepackage[backend = bibtex,
			style = numeric,
			sorting = none,
			]{biblatex}
\bibliography{references.bib}

\usepackage{url}
\usepackage{multirow}

\newcommand{\eg}[1]{}
\renewcommand{\eg}[1]{(e.g. {#1})}

\newcommand{\ie}[1]{}
\renewcommand{\ie}[1]{(i.e. {#1})}

\newcolumntype{L}[1]{>{\raggedright\let\newline\\\arraybackslash\hspace{0pt}}m{#1}}
\newcolumntype{C}[1]{>{\centering\let\newline\\\arraybackslash\hspace{0pt}}m{#1}}

\begin{document}

\title{A Comprehensive Study on Model Initialization Techniques Ensuring Efficient Federated Learning}
\author{\IEEEauthorblockN{Adwaita Janardhan Jadhav}
\IEEEauthorblockA{\textit{adwaitas28@gmail.com}
\and
\IEEEauthorblockN{Ishmeet Kaur\thanks{Both authors, Ishmeet Kaur and Adwaita Janardhan Jadhav, are currently at Apple Inc. Both authors  contributed equally to this work.}}
\IEEEauthorblockA{\textit{*Corresponding Author: ishmeet3kk@gmail.com}} 
}

}
\maketitle

\begin{abstract}

Advancement in the field of machine learning is unavoidable, but something of major concern is preserving the privacy of the users whose data is being used for training these machine learning algorithms. Federated learning(FL) has emerged as a promising paradigm for training machine learning models in a distributed and privacy-preserving manner which enables one to collaborate and train a global model without sharing local data. But starting this learning process on each device in the right way, called ``model initialization" is critical. The choice of initialization methods used for models plays a crucial role in the performance, convergence speed, communication efficiency, privacy guarantees of federated learning systems, etc. 
In this survey, we dive deeper into a comprehensive study of various ways of model initialization techniques in FL.Unlike other studies, our research meticulously compares, categorizes, and delineates the merits and demerits of each technique, examining their applicability across diverse FL scenarios. 
We highlight how factors like client variability, data non-IIDness, model caliber, security considerations, and network restrictions influence FL model outcomes and propose how strategic initialization can address and potentially rectify many such challenges. 
The motivation behind this survey is to highlight that the right start can help overcome challenges like varying data quality, security issues, and network problems. Our insights provide a foundational base for experts looking to fully utilize FL, also while understanding the complexities of model initialization.

\end{abstract}


\input{intro.tex}

\input{local.tex}

\input{central.tex}

\input{privacy}

\input{conclusions.tex}

\printbibliography

\end{document}

%% file: intro.tex
\section{Introduction}

In the era of big data and machine learning, the conventional centralized model of data processing and training is increasingly being challenged by emerging paradigms that emphasize user privacy and data security \cite{verbraeken2020survey}. One such paradigm is Federated Learning (FL) \cite{mcmahan2017communication}. 
Federated Learning is a decentralized learning approach that enables model training across multiple devices or servers while keeping data localized. Instead of transmitting raw data to a central server, devices compute model updates locally and then send these updates, ensuring user data privacy and reduced data transmission overheads. Sometimes this is even an iterative process till we receive the expected performance for the global model.
FL finds application in industries like healthcare, finance, and telecommunications, where data privacy is paramount and data-sharing can be a regulatory or logistical challenge, and in IoT devices or edge devices where resource constraint is an issue \cite{li2020review}.  
However, despite its revolutionary approach, Federated Learning comes with its own set of challenges. 

\begin{figure}[t]
\begin{center}
\centering
\subfigure[]{\includegraphics[width = 0.34\textwidth]{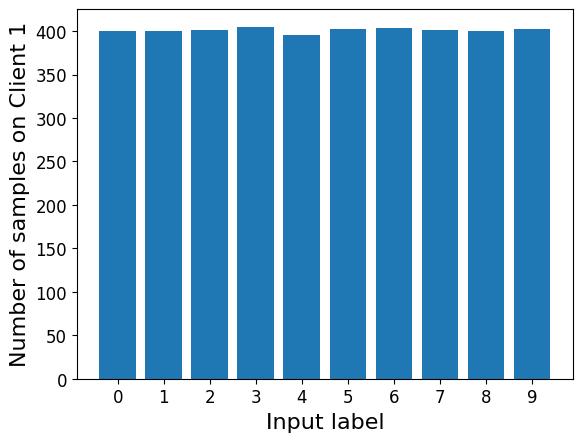}}\\
\subfigure[]{\includegraphics[width = 0.34\textwidth]{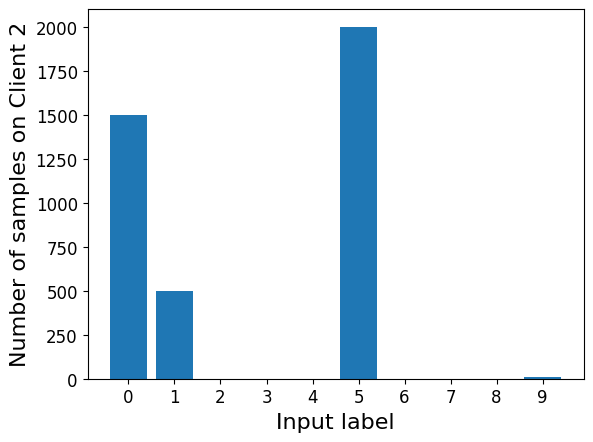}}
\end{center}
\caption{Examples of data distributions on two different clients. (a) Client 1 has a similar number of samples for all the input labels. This represents IID data. (b) Client 2 has varied numbers of samples for the input labels. This represents non-IID data.}
\label{fig:iid}

\end{figure}

\begin{enumerate}
    \item Data Non-IIDness: In FL, data is often non-independent and identically distributed (non-IID) \cite{zhao2018federated}. As you can see in Fig.~\ref{fig:iid}, This means that data across different nodes can be highly unbalanced or skewed, which can lead to models that do not generalize well across the network. 
    \item Diverse Node Characteristics: Nodes in FL can range from high-performance computing servers to everyday smartphones hence diverse node characteristics \cite{9773116}. 
    \item Communication constraints: Given the distributed nature of FL, network constraints such as latency \cite{9773116}, bandwidth limitations, and intermittent connectivity can hamper the efficient exchange of model updates. 
\end{enumerate}

Multiple efforts are being made in the direction of developing various algorithms and architectures to make Federated Learning more efficient.
Amidst these challenges, model initialization in FL emerges as a key factor determining the system's overall success\cite{nguyen2023begin}. The inspiration to dive deeper into this topic arises from the advantages of effective model initialization. In decentralized datasets, optimizing training initiation can result in quicker convergence, improved computational efficiency, better model accuracy against data non-IIDness challenges, and a balanced learning experience across varied nodes without undue data skewness from any single node ~\cite{sun2021partialfed}.
To the best of our knowledge, this is the first study where methodically  model initialization in FL is categorized and compared against FL dystem challenges and how it can benefit. 
To further make our study clear, we have categorized initialization techniques in the FL system into 3 subcategories as below:

\begin{enumerate}

    \item Localized Initialization: This focuses on leveraging device-specific data characteristics to initiate model training. These methods prioritize understanding and adapting to the unique data distribution of each participating node, promoting a more tailored and efficient learning process.
    
    \item Centralized Initialization: It utilizes a central server to initiate model training. This approach often involves pre-training a global model using vast datasets leveraging pre-existing models, which is then fine-tuned on edge devices through Federated Learning.
    
    \item Privacy-preserving Initialization: As the name suggests the emphasis is on preserving the privacy of users and avoiding data security issues while initializing the model. 

\end{enumerate}

In the following sections, we dive deeper into the above categories and compare them against the FL systems challenges , then present metrics and evaluation recommendations and in the final section, we present our research recommendations and conclude this paper.

%% file: local.tex
\section{Localized Initialization}
\label{sec:compress}
In this section, we have identified two subcategories: starting with random settings, known as Random Initialization, and tailoring the start based on specific client data, called Client-Specific Initialization.
\subsection{Random Initialization}
In the area of Federated Learning (FL), Random Initialization serves as a fundamental technique where model parameters, including weights and biases, are populated with values drawn from certain probability distributions ~\cite{10.1007/3-540-59497-3_220}. The choice of distribution is paramount as it greatly influences the model's convergence speed and performance. To list some common techniques \cite{6042874}~\cite{10.1007/3-540-59497-3_220}:
\begin{enumerate}
    \item Uniform Distribution: Parameters are sampled uniformly with random values from a given range, such as [-0.5, 0.5].
    \item Truncated Normal Distribution: This approach mirrors the normal distribution, but values beyond a determined range are excluded and replaced.
    \item Normal Distribution: Parameters are randomly drawn from a Gaussian distribution, typically with a mean of 0 and a standard deviation of 1.
\end{enumerate}

One of the intrinsic benefits of Random Initialization is that it assignes each client with distinct weights, leading to diversity during the learning process.
It also addresses pivotal challenges in FL as we discussed before, particularly Client Heterogeneity and model convergence \cite{sun2021partialfed}. By beginning with random weights, the model explores diverse regions of the parameter space, nicely tuning to each client's unique data attributes. This promotes a more adaptive learning process in a federated landscape. As random initialization provides a neutral starting point for model parameters, the model training does not make any assumptions about the data distribution leading to improvement in model convergence and avoiding poor local optima. 

Random initialization, while useful, is not without its limitations. As indicated by ~\cite{nguyen2023begin}, certain initialization techniques can surpass the performance of random initialization, with evaluations drawing from diverse datasets like CIFAR-10, FEMNIST, Stack Overflow, and Reddit. Moreover, random initialization's impact on communication efficiency is noteworthy. It may also result in initial models with large parameter differences across clients. During the training process, when model updates are aggregated at the central server, these pronounced differences could translate to heftier model update transmissions, thereby amplifying the communication overhead during aggregation.

\subsection{Client-Specific Initialization}

Client-specific initialization in federated learning tailors the initialization of local models based on the unique data attributes of individual clients. There are scenarios where recognizing the distinctiveness of each edge device becomes important. For instance, personal autocomplete tasks like \textit{"I love dancing"} can be particular to an individual. While models drawing from expansive global data shows enhanced performance, those focused on localized, client-specific data offer the benefit of capturing individual data characteristics.

The Client-Specific initialization techniques can be further divided into the following categories. 

\begin{enumerate}
    \item Data-Driven Initialization: Prior to model initialization, every client conducts an analysis of their local data, identifying statistical properties like data distribution, imbalances, and distinct patterns. This process helps to highlight properties unique to that client, which are then used to set the model's weights and biases.

    \item Domain-Specific Initialization: A model trained for a particular domain may not be directly useful for another domain's dataset, even though there might be some common knowledge between these domains.\cite{NEURIPS2021_c429429b} refers to the practice of initializing the model parameters of a federated learning model using domain-specific knowledge or pre-trained models that are relevant to the target task in each participating device (client) before starting the federated learning process. This helps to improve convergence and overall performance by providing a good starting point for the optimization process. For example, sentiment analysis for tweets, newspapers, and hospital reviews is very different, but they might have some commonality. 
    
    \item Clustering-Based Initialization:  Cluster-based initialization groups client devices by their similarities, then tunes local models based on client-specific attributes before the federated learning commences.\cite{10.1145/3558005}.For instance, in diverse Natural Language Processing tasks, data might range from news articles to social media posts and song lyrics. By categorizing clients with like data and adjusting the local model through averaging or other aggregation techniques, one can optimize for that cluster's specificities. While this method enhances privacy, it's vulnerable to byzantine clients \cite{shejwalkar2021manipulating}, where a malicious initialization could skew the global model's training and yield inaccurate results.
\end{enumerate}
In FL systems, client-specific initialization offers notable advantages. Firstly, it addresses model fairness and bias, promoting better equity by tailoring models to each client's personalized data. Secondly, it mitigates the challenges posed by non-IID data. By initializing models that resonate with their local data distributions, clients can begin with models predisposed towards their specific data, enhancing convergence for their local tasks. This method also fosters greater personalization, marking it as a compelling area for ongoing research. However, challenges persist, notably the risk of byzantine clients \cite{portnoy2021federated} who may malevolently initialize models, potentially undermining the global training process.

%% file: central.tex
\begin{figure*}[t]
\begin{center}
\includegraphics[width=0.63\linewidth]{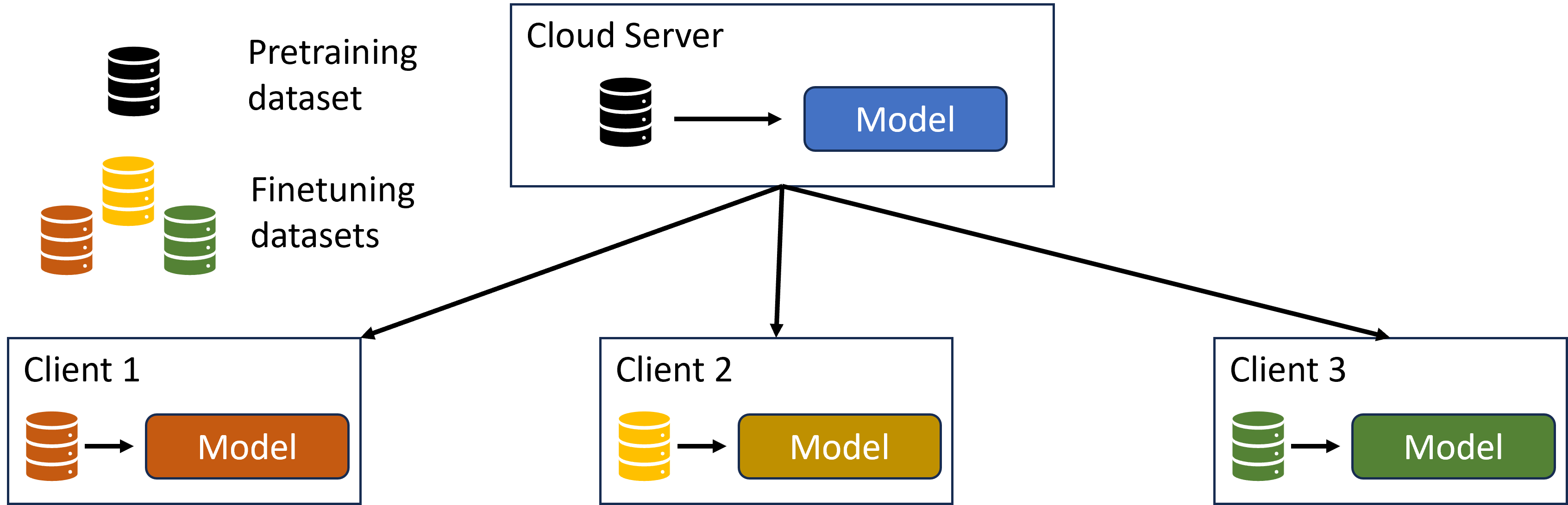}
\end{center}
\caption{Example of pre-training initialization on the federated learning cloud server. Once the model is pre-trained, the model is used to initialize the local models on each client. Each client updates its local model's weights to fine-tune the model.}

\label{fig:pretrain}
\end{figure*}

\section{Centralized Initialization}

Centralized Pre-training Initializationis the technique in which the global model is first initialized on the global server and then distributed to the local servers. This type of technique is depicted in Fig.~\ref{fig:pretrain}. The global model initialization uses two techniques as Pre-training on the Central Server and Transfer Learning.
This method aims to leverage the benefits of pre-training, which helps the model learn general features from a large dataset, while also enabling personalized fine-tuning on decentralized data sources.

\subsection{Pre-training on Central Server}
Given the vast data available online, models can be pre-trained on these large datasets prior to distribution to edge devices. Research\cite{chen2023importance} indicates that such pre-training enhances task accuracy and addresses federated learning challenges, notably non-IID client data.
Pre-training requires extensive and varied datasets. If unavailable, synthetic data can be used. The model is initially trained on a central server with substantial computational power, learning general features, semantic relationships, and linguistic structures present in the data.This process typically involves unsupervised learning methods like language modeling, autoencoders, or other self-supervised learning techniques. Tasks like image classification have notably benefited from this approach \cite{chen2022pre}.
Pre-training a model on a centralized dataset offers several benefits for FL. Firstly, it improves generalization as the model learns general features and patterns. This ensures the model's adaptability to varying data distributions across individual devices. Additionally, initializing a model through pre-training on a resource-rich central server can accelerate convergence during device-specific fine-tuning. However, this approach has its challenges. For instance, the complexity of fine-tuning rises post pre-training. Depending on the model's architecture, careful parameter adjustments and regularization might be necessary to prevent overfitting during the local device training.

\subsection{Transfer Learning}

Transfer-learning-based initialization \cite{raghu2019transfusion} is a technique where a pre-trained model's knowledge is used, and the model's parameters such as weights and bias are used as an initial starting point for training a new model on a target task. Transfer learning \cite{torrey2010transfer} is used when the knowledge might be sharable between domains, or there might not be enough data available for the target task.
Implementing transfer learning in federated learning offers great benefits. Firstly, by using pre-trained weights, models achieve quicker convergence, by utilizing the prior knowledge from the parameters. This rapid training ensures efficient data usage. Secondly, the generic features achieved by pre-trained models enhance their generalization capabilities, enabling them to perform well even with limited data specific to the target task. However, challenges still exist. The efficiency of transfer learning depends on the likeness between source and target domains; vast dissimilarities might render it ineffective \cite{chen2020fedhealth}. Moreover, the relevancy of tasks is crucial: non-aligned tasks can make the transferred knowledge redundant. Careful fine-tuning is essential to avoid overfitting and potential biases or noise from the source data, which might adversely impact the target task's performance.

%% file: privacy.tex
\section{Privacy-preserving Initialization}
\label{sec:design}
With the rise of various applications of Machine Learning, privacy is one of the major concerns of most of the governments of various countries, The European Union even passed various laws such as GDPR \cite{basin2018purpose}. Although one of the major advantages of Federated Learning is said to be Privacy preservation, few studies have found that hackers and attackers have the potential to deduce the specifics from local models exchanged between clients \cite{DBLP:journals/corr/McMahanMRA16}. There are three major types of privacy-preserving initialization which are Homomorphic Encryption-based initialization, Secure Multi-Party Computation (SMPC) Initialization, and Differential Privacy-based Initialization. SMPC and Homomorphic Encryption are Encryption-based techniques whereas in Differential privacy initialization noise is added at the time of model initialization.

\subsection{Homomorphic Encryption-based Initialization}

Homomorphic Encryption-based Initialization \cite{9812492} safeguards client data privacy during the global model's initiation.  It uses homomorphic encryption techniques which allows the computations to be performed on encrypted data without the need to decrypt it.

Homomorphic encryption is a lattice-based public key system \cite{melchor2008lattice} that permits operations on encrypted text without decryption. Using a public key for encryption and a secret key for decryption, the sent message represents the input's sum and product, which is deciphered using the secret key. 

The technique is further categorized as :
 \begin{enumerate}
\item Partially Homomorphic Encryption: This method uses only a single operation either addition or subtraction. Some examples of Partially Homomorphic Encryption are 
ElGamal encryption \cite{7231798} which utilizes the multiplication operation and Paillier encryption\cite{7231798} which utilizes the addition operation.

\item Somewhat Homomorphic Encryption: This type supports both addition and multiplication operations on encrypted data, but if overdone then the text might be susceptible to security issues. The RSA encryption scheme \cite{aboud2008efficient} is an example of somewhat homomorphic encryption.

\item Fully Homomorphic Encryption (FHE): Despite repeated operations, encryption remains intact without needing decryption. FHE schemes \cite{gentry2009fully} are resource-intensive and less practical, they are potent for privacy applications, such as the BGV \cite{gentry2012ring} and TFHE methods \cite{chillotti2020tfhe}.

\end{enumerate}

Homomorphic encryption is protected from quantum threats due to its lattice-based foundation. Furthermore, it offers adjustable security parameters, guaranteeing its safety under standard assumptions, and bolstering its defense against adversaries with limited computational capabilities.
Homomorphic encryption-based initialization in federated learning setup involves initializing and updating the parameters of the local model with the decrypted global model data. This ensures privacy, security, and client autonomy for local updates with encrypted data. However, there are computational and communication challenges due to encryption operations. We suggest combining this technique with client-specific initialization to boost security and privacy.

\subsection{Secure Multi-Party Computation (SMPC) Initialization} 
Secure Multi-Party Computation Initialization \cite{goldreich1998secure} initializes the global model without revealing client details, allowing computations on encrypted data where only the final results are disclosed.

Steps involved in SMPC for FL \cite{byrd2020differentially}:
\begin{enumerate}
    \item Local Data Encryption: Every client can encrypt their weights and send a part of it to other clients.
\item Collaborative Computation: The client performs computations on the encrypted data, and these computed values are communicated to the global server. They evaluate mathematical functions over their encrypted inputs without exposing the individual data.
\item The global server can then perform aggregation on these values to find the final answer.
\end{enumerate}

SMPC ensures privacy and client data confidentiality during initialization but faces challenges like computational overhead from encryption and algorithm complexity, which may slow convergence. We recommend its use, especially prioritizing security, combined with client-specific initialization for personalized and private results.

\begin{table*}[t!]
    \centering
    \caption{Comparison of Model Initialization Techniques in Federated Learning}
    \begin{tabularx}{\textwidth}{lXXX}
        \toprule
        \textbf{Feature/Technique} & \textbf{Localized Initialization} & \textbf{Centralized Initialization} & \textbf{Privacy-preserving Initialization} \\
        \midrule
        Main Concept & Device-specific data characteristics. & Global model on central server. &  User data privacy during initialization. \\
        \midrule
        Key Methods & Device-based tailoring. \newline Data-specific adaptation. & Cloud-based pre-training. \newline Global-to-local fine-tuning. & Differential privacy. \newline Encrypted computation. \\
        \midrule
        Benefits & Tailored learning experience. \newline Better local accuracy. & Rapid start with pre-trained model. \newline Consistency in initial model across devices. & Enhanced user trust. \newline Regulatory compliance. \\
        \midrule
        Challenges & Varied starting points can complicate global model aggregation. & Requires significant server resources. \newline May not reflect local data nuances. & Computational overhead. \newline Possible accuracy trade-off. \\
        \midrule
        Convergence Speed (relative). & Moderate. & Fast. & Slow. \\
        \midrule
        Computational Efficiency(relative). & High. & Medium. & Low due to encryption processes. \\
        \midrule
        Suitability for Data Non-IIDness & Very Suitable (as it tailors to local data). & Moderate. & Moderate. \\
        \midrule
        Privacy \& Security Measures. & Local data stays local. & Centralized data aggregation. & Data encryption. \newline Noise addition for privacy. \\
        \midrule
        Applicability (Use Cases) & Edge devices with diverse data. \newline IoT devices. & Situations with robust central servers. \newline Scenarios where pre-trained models are available. & Healthcare and finance where data privacy is paramount. \newline User-centric applications. \\
        \midrule
        Drawbacks & Might introduce inconsistency in global model. & Might overlook unique local data patterns. & Slower due to privacy processes. \newline Might introduce noise to data. \\
        \bottomrule
    \end{tabularx}
    \label{tab:one}
\end{table*}

\subsection{Differential Privacy-based Initialization}

Differential Privacy-based initialization \cite{wei2020federated} protects individual data privacy without impacting the model's results. It conceals individual specifics by assessing each data point's influence on the model and adding noise during initialization. This ensures data identities remain hidden and prevents encoding of sensitive information in initial parameters.

Four key steps in DP-based initialization are:

\begin{enumerate}
 \item Sensitivity Computation: Here, the model's sensitivity to changes in local node initialization parameters is evaluated. It measures the shift in model predictions when a single data point enters or leaves the training dataset.

\item Noise Addition: After calculating the sensitivity, some noise is added to the model's parameters during initialization. The addition of this noise makes it more secure since there are very less chances for it to carry any specific information from the training data.

\item Privacy Budget:Represented by epsilon, this metric quantifies the privacy protection level. A lower epsilon means added noise for enhanced privacy assurances.

\item Trade-off: There's a trade-off between privacy and utility. Stronger privacy protection usually implies more noise, which can degrade the model's performance. Balancing this trade-off is a crucial consideration in Differential Privacy-based Initialization. \cite{app13074168}
\end{enumerate}

In federated learning emphasizing DP, clients add noise to model parameters before sending them. This communication's cost aligns with standard federated learning. Unlike the two aforementioned encryption-based initialization methods, DP offers reduced computation and communication burdens. Future enhancements should retain client result personalization while preserving privacy.

%% file: conclusions.tex
\section{Discussion and Proposed Evaluation Metrics}

\subsection{Findings and Discussion}

In our exploration of model initialization techniques in Federated Learning (FL), we discerned three primary approaches: Localized, Centralized, and Privacy-preserving Initialization. We summarize our findings in TABLE~\ref{tab:one}. Localized Initialization, tailored to individual devices, excels in environments with diverse data distributions but may challenge global model consistency. Centralized Initialization offers speed and uniformity but could gloss over local data nuances. Meanwhile, Privacy-preserving Initialization prioritizes user data security, vital in sensitive sectors like healthcare and finance, but introduces computational challenges and potential trade-offs in model accuracy. While some research has delved into FL's intricacies, our systematic categorization and comparison focused on model initialization provide an essential roadmap for researchers and practitioners, underscoring the nuances and critical considerations previously not noted in the realm of Federated Learning. 

\subsection{Proposed Evaluation Metrics}

In evaluating the efficacy of model initialization techniques in Federated Learning (FL), the metrics play a pivotal role in understanding their performance under varying conditions and system settings \cite{lai2022fedscale}. Below are some metrics criteria we introduce and recommend to choose the best initialization strategy based on the discussed challenges of FL system.
\subsubsection{Quantitative Metrics}
\begin{itemize}
    \item Convergence Speed: It's about the number of times data needs to be exchanged between the server and nodes before the model starts giving good results and meeting certain accuracy benchmarks.Faster convergence means less time and potentially less resource consumption.
    \item Communication Overhead: Quantifying the amount of data exchanged between the nodes and the central server and the time taken during initialization. In scenarios where bandwidth is limited or costly, minimizing communication is vital \cite{da2023resource}.
    \item Computational Load: This assesses how much computer power is needed in the early stages of model training. If the computational load is too high, it might exclude devices with less processing power or lead to longer training times.
\end{itemize}

\subsubsection{Qualitative Metrics }
\begin{itemize}
    \item Scalability: Evaluate the ability of the initialization technique to cater to an increasing number of nodes or devices. One can run multiple experiments with varying numbers of nodes (e.g., 10, 100, 1000) and observe if the initialization technique maintains efficiency and effectiveness.
    \item Robustness: Assesses the initialization technique's performance under real-world challenges like device dropouts, asynchronous updates, etc. 
    \item Interoperability: Measures the compatibility of the initialization method with various FL architectures and algorithms.
\end{itemize}

\section{Conclusion}
Federated learning systems are filled with challenges that are further made complex when deciding on the most optimal model initialization technique. Through our comprehensive survey, it becomes evident that the right initialization can significantly boost the performance, speed, and robustness of FL models. We have compared and categorized some major challenges of FL systems and propose potential solutions for them using various initialization techniques. We also introduced a new categorization and metrics for researchers to navigate Federated Learning Systems. As the usage of decentralized learning continues to expand, so will the techniques to initialize the process, ensuring that Federated Learning maintains its status of privacy-preserving and resource-efficient machine learning technique.